Title: The Case Records of ChatGPT: Language Models and Complex Clinical Questions

Authors:
- Timothy Poterucha, MD[a]
- Pierre Elias, MD[a,b,c]
- Christopher M. Haggerty, PhD[b,c]

Affiliations:
[a]Department of Medicine, Columbia University Irving Medical Center/NewYork-Presbyterian Hospital, New York, New York United States of America
[b]Department of Biomedical Informatics, Columbia University, New York, New York United States of America
[c]NewYork-Presbyterian Hospital, New York, New York United States of America

**Address for Corresponding Author:**
Timothy Poterucha, MD
177 Fort Washington Ave, Milstein Hospital
New York, NY 10032
Email: tp2558@cumc.columbia.edu

Disclosures: TJP owns stock in Abbott Laboratories and Baxter International with research support provided to his institution from the Amyloidosis Foundation, American Heart Association, Eidos Therapeutics, Pfizer, Edwards Lifesciences, and the New York Academy of Medicine.

Word count: 625

**Abstract**

**Background**: Artificial intelligence language models have shown promise in various applications, including assisting with clinical decision-making as demonstrated by strong performance of large language models on medical licensure exams. However, their ability to solve complex, open-ended cases, which may be representative of clinical practice, remains unexplored.

**Methods**: In this study, the accuracy of large language AI models GPT4 and GPT3.5 in diagnosing complex clinical cases was investigated using published Case Records of the Massachusetts General Hospital. A total of 50 cases requiring a diagnosis and diagnostic test published from January 1, 2022 to April 16, 2022 were identified. For each case, models were given a prompt requesting the top three specific diagnoses and associated diagnostic tests, followed by case text, labs, and figure legends. Model outputs were assessed in comparison to the final clinical diagnosis and whether the model-predicted test would result in a correct diagnosis.

**Results**: GPT4 and GPT3.5 accurately provided the correct diagnosis in 26% and 22% of cases in one attempt, and 46% and 42% within three attempts, respectively. GPT4 and GPT3.5 provided a correct essential diagnostic test in 28% and 24% of cases in one attempt, and 44% and 50% within three attempts, respectively. No significant differences were found between the two models, and multiple trials with identical prompts using the GPT3.5 model provided similar results.

**Conclusions**: In summary, these models demonstrate potential usefulness in generating differential diagnoses but remain limited in their ability to provide a single unifying diagnosis in complex, open-ended cases. Future research should focus on evaluating model performance in larger datasets of open-ended clinical challenges and exploring potential human-AI collaboration strategies to enhance clinical decision-making.

**Manuscript**

**Background and Objectives**

Artificial intelligence language models have shown promise in various applications, including assisting with clinical decision-making. Prior work has identified high model performance in answering multiple choice, narrow-scope medical licensing questions (1). In contrast, clinical medicine often involves open-ended, complex cases. We aimed to study the accuracy of large language models in the diagnosis of real-world, complex clinical cases using published Case Records of the Massachusetts General Hospital.

**Methods**

Case Records of the Massachusetts General Hospital that were published between January 1, 2022 and April 16, 2023 were accessed to identify a total of 50 cases that called for a diagnosis and diagnostic test to be performed. The ability of the large language AI models GPT4 and GPT3.5 (OpenAI, California) were tested in their ability to correctly identify the correct diagnosis and diagnostic test. The dates were selected to avoid data contained in model training datasets (2).

Models were given a prompt requesting the top three specific diagnoses and associated diagnostic tests, followed by case text, labs, and figure legends. The final prompt provided to the model for this analysis was the following: "This is a clinical case series. In bullet point form, I would like you to tell me your top three specific diagnoses (in order of likelihood) with the associated essential diagnostic test that should be performed. Do not give reasoning, just the diagnosis and test with a period between them. These may be rare cases, and the diagnosis should be specific rather than general (for instance, iron deficiency anemia due to GI bleeding rather than just anemia)". All analyses used the March 23, 2023 version of ChatGPT.

**Results**

GPT4 identified the correct diagnosis on the first choice in 13 cases (26%), while GPT3.5 did so in 11 cases (22%) (Figure 1A). For top three choices, GPT4 made the correct diagnosis in 23 cases (46%) and GPT3.5 in 21 cases (42%). GPT4 correctly identified the essential test on the first try in 14 cases (28%) and GPT3.5 in 12 cases (24%) (Figure 1B). Within the top three choices, GPT4 identified the correct test in 22 cases (44%) and GPT3.5 in 25 cases (50%). Chi-squared testing in Stata 17.0 revealed no significant differences between GPT4 and GPT3.5 ($p > 0.05$ for all comparisons). In some cases, GPT4 provided a more specific diagnosis than GPT3.5, such as Cotard's syndrome for Case 34-2022 and pembrolizumab-induced diabetes in Case 6-2022 (3, 4). Repetitive trials were conducted in which the GPT3.5 model was given the same prompt an additional 4 times for a total of 5 trials for all 50 tested cases. In the 39 cases where the first diagnosis was incorrect on Trial 1, the first diagnosis was correct in an additional 2 cases on one of the subsequent 4 trials. When all 3 diagnoses from all 5 trials were included, the correct diagnosis was reached in 31 of the 50 cases (62%).

**Discussion**

This analysis demonstrates the application of large language models in complex clinical cases, with GPT4 and GPT3.5 providing an accurate diagnosis in 22-26% of cases in one and 42-46% within three attempts. Thus, these models may be useful for generating differential diagnoses but frequently fail to provide a unifying diagnosis. Although complex, the open-ended nature of these cases may more closely mirror general medical practice where patients frequently have extensive histories, comorbidities, and multiple testing abnormalities, in contrast to standardized medical licensure testing. This work highlights that, in their current state, these may be helpful tools but remain limited compared to expert clinicians. Limitations of the study include small available sample size, single prompt type, and lack of a human comparator. Future studies are needed to test model performance in larger datasets of open-ended clinical challenges and explore human-AI collaboration strategies.

**Figure 1A and 1B: Comparison Between GPT4 and GPT3.5 in Providing the Correct Diagnosis (Figure 1A) and Correct Diagnostic Test (Figure 1B) in Case Records**

Legend: The language models GPT4 and GPT3.5 were provided the text from 50 cases from the Case Records of the Massachusetts General Hospital and prompted to provide the three most likely diagnoses and essential diagnostic tests. Both models were able to identify the correct diagnosis or diagnostic test 22-28% of the time within one guess and 42-50% of the time within three guesses with no statistically significance difference between the two models.

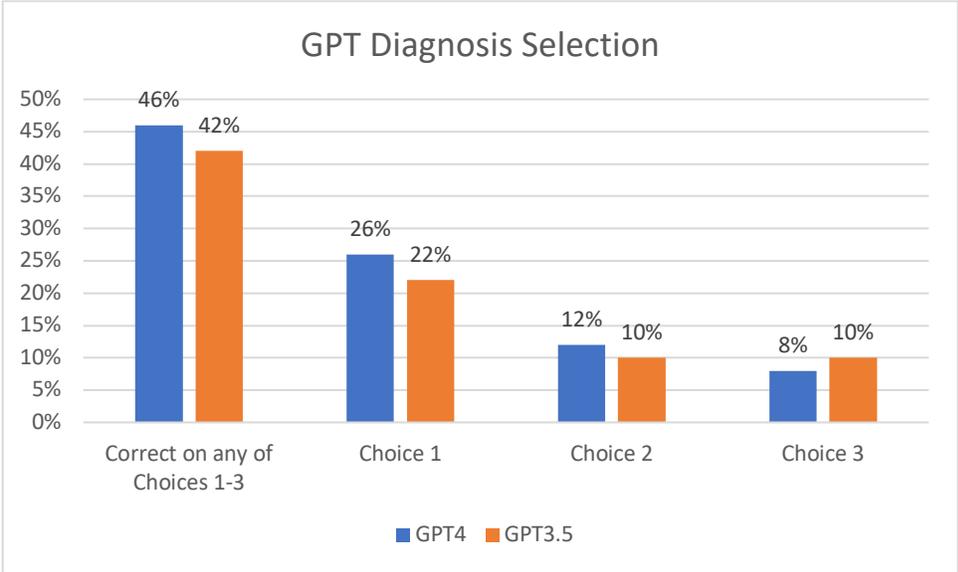

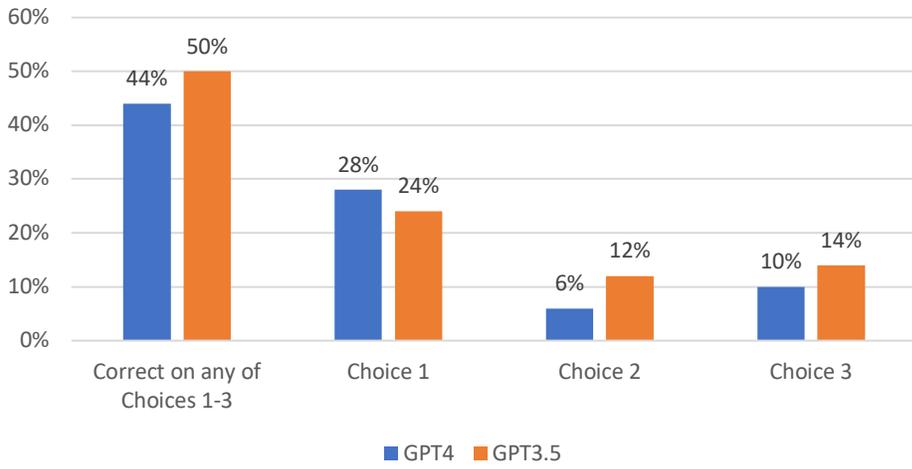